\documentclass[11pt]{article}

\usepackage[final]{acl}

\usepackage{times}
\usepackage{latexsym}

\usepackage[T1]{fontenc}

\usepackage[utf8]{inputenc}

\usepackage{microtype}

\usepackage{inconsolata}

\usepackage{graphicx}
\usepackage{multirow}
\usepackage{booktabs}
\usepackage{tcolorbox}
\usepackage{hyperref}
\usepackage{algorithm}
\usepackage{algpseudocode}
\usepackage{amsmath}
\usepackage{tabularx}

\usepackage[normalem]{ulem}

\title{Trification: A Comprehensive Tree-based Strategy Planner and Structural Verification for Fact-Checking}

\author{
 \textbf{Anab Maulana Barik}\thanks{\,This work was done during an internship at Huawei Celia Team.},
 \textbf{Shou Ziyi},
 \textbf{Yang Kaiwen},
 \textbf{Yang Qi},
 \textbf{Shen Xin\thanks{\,Corresponding author} }
\\
 Huawei Celia Team\\
\\
 \texttt{
   maulanaanab@gmail.com, shenxin19@huawei.com
 }
}

\begin{document}
\maketitle
\begin{abstract}
Technological advancement allows information to be shared in just a single click, which has enabled the rapid spread of false information.
This makes automated fact-checking system necessary to ensure the safety and integrity of our online media ecosystem.
Previous methods have demonstrated the effectiveness of decomposing the claim into simpler sub-tasks and utilizing LLM-based multi agent system to execute them.
However, those models faces two limitations: they often fail to verify every component in the claim and lack of structured framework to logically connect the results of sub-tasks for a final prediction.
In this work, we propose a novel automated fact-checking framework called \texttt{Trification}.
Our framework begins by generating a comprehensive set of verification actions to ensure complete coverage of the claim.
It then structured these actions into a dependency graph to model the logical interaction between actions.
Furthermore, the graph can be dynamically modified, allowing the system to adapt its verification strategy.
Experimental results on two challenging benchmarks demonstrate that our framework significantly enhances fact-checking accuracy, thereby advancing current state-of-the-art in automated fact-checking system.
\end{abstract}

\section{Introduction}
Technological advancements have fundamentally changed how online media operates.
The rise of internet and social media platforms allow information to be shared with just a single click.
However, this ease of sharing has also enabled the rapid spread of false information, underscoring the critical need for automated fact-checking system.

Automated fact-checking systems have evolved significantly over the past decades. 
Early approached \cite{ukp-athene,gear,kgat,dream} typically followed a standard three-step pipelines: (1) document retrieval for retrieving relevant document, (2) sentence selection for extracting top-k candidate evidence, and (3) veracity prediction for label prediction.
With the recent advancements of Large Language Models (LLMs), the paradigm is now shifting towards more integrated, LLM-based agent methods.
Instead of executing predefined steps, LLM-based agents leverages the LLM's reasoning capabilities to plan a verification strategy, interact with external tools, and synthesize information to arrive at a final verdict. 
These strategies have primarily focused on steps like claim decompositions, retrieval method, question answering, and final reasoning \cite{pan2023factchecking, zhao-etal-2024-pacar, local, search-in-the-chain}.

\begin{figure*}[t]
    \centering
    \includegraphics[width=\linewidth]{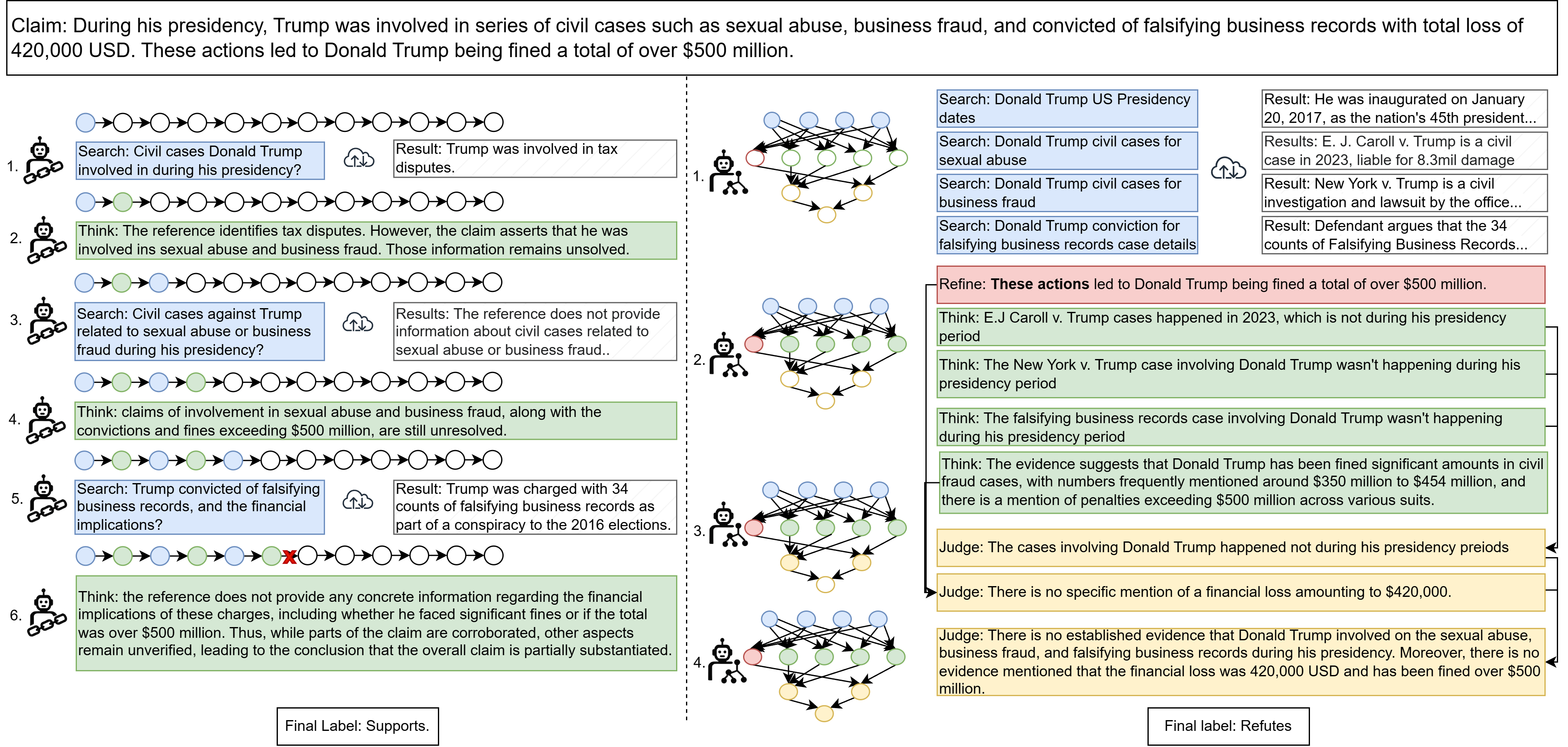}
    \caption{Case Comparison: Chain Search vs. Tree Verification}
    \label{fig:case-comparison}
\end{figure*}

Despite the success of these LLM-based fact-checking agents, they often suffer two critical shortcomings, as shown in \autoref{fig:case-comparison}.
First, they frequently fail to perform complete verification on every component within a complex claim, often because of the dependence on a linear reasoning path that may overlook short or implicit statements in the claim.
Second, they lack a mechanism to make logical connections between the results of each sub-tasks because they often treat them as isolated steps rather than interdependent nodes in a reasoning graph.
This prevents the agents to perform logical and more comprehensive reasoning, leads to logical errors and causal issues \cite{local}.

Therefore, to address the above limitations, we propose a fact-checking framework that leverages a tree-based strategy planner and structural verification process called \texttt{Trification}.
Unlike earlier systems that generate sub-tasks sequentially \cite{local,zhao-etal-2024-pacar, search-in-the-chain} -- a process prone to premature termination, our framework begins by generating a complete set of verification actions.
This approach ensures that every part of the claim is covered from the set.
Our framework then structures these actions into a dependency graph, which provides two key advantages.
First, in contrast to previous methods that combine all sub-tasks' results at the final step \cite{pan2023factchecking, zhao-etal-2024-pacar}, our graph introduces intermediate logical verification between dependent actions.
This enables a more fine-grained and rigorous evaluation of a claim.
Second, by explicitly modeling dependencies, ready-to-execute actions (those with no pending dependencies) can be processed concurrently, alleviating the computational bottleneck inherent in a sequential process.

Moreover, to prevent our system to end prematurely or produce non-sensical output, our framework allows for dynamic modification of the verification graph when a specific action fails.
In such cases, the system is prompted to generate a new sub-graph with an alternative verification strategy, which is then integrated into the original verification graph.
This iterative recovery mechanism enhances the systems adaptability, mirrors the creative approach of human fact-checkers.

In summary, our main contributions are
\begin{enumerate}
    \item We propose \texttt{Trification}, a novel fact-checking framework that uses a tree-based strategy to generate a comprehensive set of verification actions, to ensure no component of a claim is overlooked.
    \item We introduce an executable dependency graph to structure the verification actions. This graph explicitly models logical relationships between actions, enabling more fine-grained reasoning and concurrent execution to reduce computational bottlenecks.
    \item We demonstrate a graph modification mechanism that allows the system to adapt its strategy upon failing, enhancing the creativity to explore alternative verification strategy.
\end{enumerate}

\section{Related Work} 


Automated fact-checking has undergone substantial evolution, shifting from modular pipelines to LLM-based reasoning agents.
Early pipeline systems, such as UKP-Athene \cite{ukp-athene}, laid the foundation by combining document retrieval with neural entailment models for claim classification. However, these systems typically processed evidence in isolation, limiting their ability to capture inter-evidence dependencies.
To address this, subsequent work introduced graph-based reasoning architectures that modeled relationships among multiple evidence pieces. GEAR \cite{gear} employed a fully connected evidence graph with BERT-based aggregators to enable information transfer between nodes, while KGAT \cite{kgat} refined this design using kernel-based attention to control evidence propagation and assess node importance more precisely. Later systems extended this idea to semantic-level graphs, leveraging semantic role labeling and Graph Neural Networks over contextual encoders to reason over structured evidence representations \cite{dream}. 
This entire evolution fundamentally underscores the critical importance of structured evidence processing for robust fact-checking but limited by their reliance on supervised learning strategies. 

Recent advances in Large Language Models have enabled a new generation of agent-based fact-checking systems, where the model plans and executes verification steps. 
Instead of adhering to a fixed pipeline, LLM-based agents leverage their reasoning abilities to decompose claims, retrieve evidence, and synthesize information across multiple sources \cite{pan2023factchecking, zhao-etal-2024-pacar, local, search-in-the-chain}. This paradigm shift has improved flexibility and interpretability, allowing models to mimic human-like investigative workflows, however, they still face two major limitations.
First, their reasoning often follows a linear verification chain, which can overlook implicit or short sub-claims embedded within complex statements. 
Second, existing agents generally treat subtasks as independent modules, lacking a coherent framework for connecting and reasoning over the interdependencies among sub-claims or evidence. Without such structured reasoning, models are prone to logical inconsistencies and causal misinterpretations, as observed in recent analyses \cite{local}.
In contrast to sequential approaches, 
our tree-based fact-checking framework introduces a structured and adaptive reasoning process.
It enables the concurrent execution of independent tasks and incorporating dynamic graph modification to ensure robust and adaptive reasoning.

\section{Methods}
Given a claim $C$, the goal of fact-checking is to assess its accuracy by assigning a label $y$, either \textit{supports or refutes}, indicating the veracity label of the given claim. The process involves gathering a set of evidence $E$ from the provided knowledge base or the internet. 

\subsection{Tree Planner Generation}
To effectively address these challenges and improve operational efficiency—particularly in parallel processes—we introduce \texttt{Trification}, a comprehensive tree-based strategy planner and structural verification framework.
Our approach models the fact-checking process as a dependency graph, utilizing a tree planner to establish a clear and organized verification workflow. This design underpins the core of our \textbf{Tree-based Strategy Planner} and \textbf{Structural Verification system}, known as \texttt{Trification}.

Trification generate an initial verification plan based on the input claim.
This initial plan is represented as a directed acyclic graph (DAG), where nodes correspond to specific verification actions and edges define their dependencies.
The DAG structure facilitates concurrent action execution, significantly improving computation efficiency by parallel processing. 

The framework starts by constructing an initial Directed Acyclic Graph $(\mathcal{G})$, where nodes representing verification actions, and a set of edges defining dependencies between them. 
The framework employs four node types, each representing a specific action in the verification process: (1) \texttt{SEARCH} Node to perform evidence retrieval, (2) \texttt{REFINE} Node to resolve ambiguity in the node's input, (3) \texttt{THINK} Node to perform intermediate logical reasoning by synthesizing information from multiple nodes, and (4) \texttt{JUDGE} Node to predict the final verdict.
Once the graph $\mathcal{G}$ is generated, the framework executes the actions concurrently, as long as no dependencies are blocking their execution. 

\begin{figure*}
    \centering
\includegraphics[width=\linewidth]{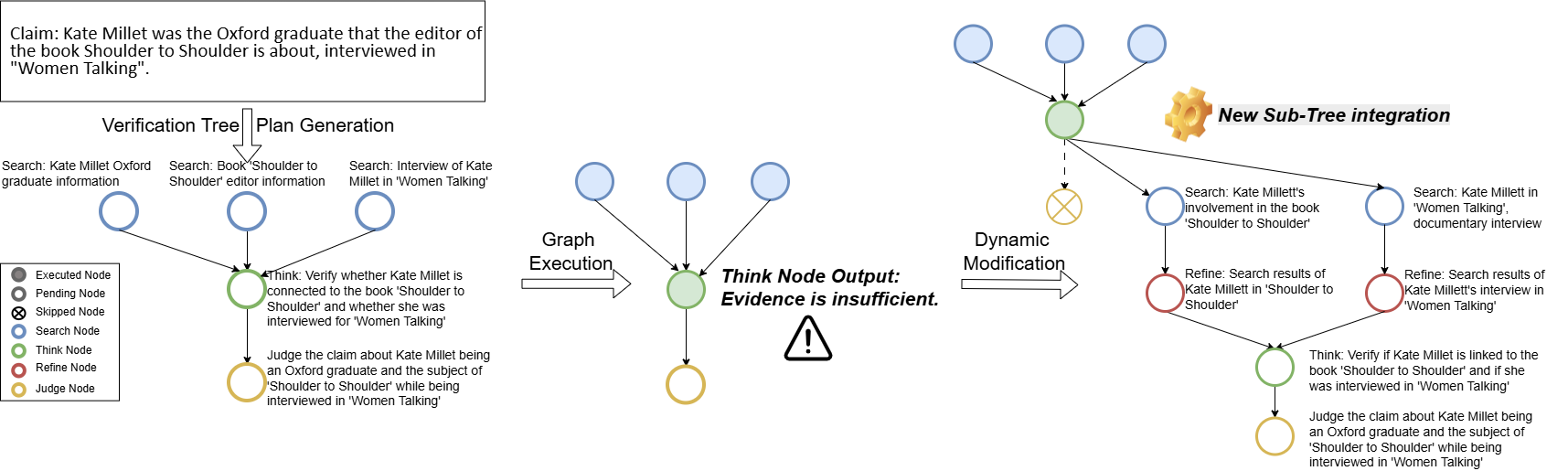}
    \caption{Dynamic Tree Planner: insufficient evidence triggers dynamic graph modification and integration of a newly generated subtree}
    \label{fig:dynamic-tree}
\end{figure*}
\subsection{Node Specification}
Each node in the graph is an object characterized by the following mandatory attributes:
\begin{itemize}
\item \textbf{id}: The node's unique identifier.
\item \textbf{type}: The type of node, which determines its operations. The four possible types are \texttt{SEARCH}, \texttt{REFINE}, \texttt{THINK}, and \texttt{JUDGE}.
\item \textbf{input}: The specific instruction or task the node is designed to execute.
\item \textbf{hint}: The additional context that defines both the integration logic for parent node outputs and the node's role within the global verification strategy.
\item \textbf{dependencies}: A list of ids corresponding to the node's parent nodes. A node is eligible for execution only when all its dependencies have been successfully processed.
\end{itemize}

The specifications for each node type are detailed in the following subsections.

\paragraph{\texttt{SEARCH} Node}
The \texttt{SEARCH} node is responsible for executing a search operation to retrieve relevant information for subsequent steps in the process. Its input is a query string, which represents the specific information or evidence needed to validate the claim or address the question at hand.

The node begins by utilizing an LLM to generate a query search based on the node's input and hints.
The query search is conditioned on the node's input and strategic hints, ensuring it retrieves information relevant to downstream nodes in the verification graph.

For each generated query search, we perform evidence retrieval using one of the two strategies: Wiki or Search Engine.
The Wiki strategy follows ProgramFC \cite{pan2023factchecking}, using BM25 to retrieve top-k paragraphs from the Wikipedia dump.
The Search Engine strategy employs 
the open-source search tool Dux Distributed Global Search (DDGS)\footnote{DDGS: \href{https://github.com/deedy5/ddgs}{https://github.com/deedy5/ddgs}}, passing the query search and collecting the top-k returned snippets.

The search results are then stored in the evidence field as a list of pertinent data points. The \texttt{SEARCH} node synthesizes the gathered information, ensuring that it aligns with the original query to provide meaningful and useful results for the subsequent nodes in the process.


\paragraph{\texttt{THINK} Node}
The \texttt{THINK} node functions as a central reasoning component that processes a specific declaration requiring verification by first synthesizing all available evidence from its parent dependencies. 
It then engages an LLM to perform intermediate reasoning over this synthesized information, aiming to produce a coherent conclusion.

\paragraph{\texttt{REFINE} Node}
The \texttt{REFINE} node acts as a critical pre-processing agent designed to resolve ambiguity in the original user query.
For instance, given the claim in \autoref{fig:case-comparison} , a node might be generated with the following input: ``These actions led to Trump being fined over \$500 million''.
This node would replace the ambiguous phrase ``These actions'' with the specific case names, which is essential for proper verification.

\texttt{REFINE} node resolves ambiguity by leveraging contextual evidence gathered from its dependency nodes. 
The process begins by collecting the outputs from all its parent nodes, which typically contain relevant search results or established facts. It then employs an LLM to analyze this contextual information to identify and resolve vague references—such as ambiguous pronouns like "they" or "those cases". The core function of the node is not to evaluate the truth of a claim but to generate a new, precise, and disambiguated query that can be effectively processed by subsequent operational nodes. The output is an optimized query text, which subsequently replaces the original input for downstream nodes, ensuring that those nodes will operate on a clear and unambiguous foundation.

\paragraph{\texttt{JUDGE} Node}
The \texttt{JUDGE} node serves as the terminal verdict mechanism within the reasoning graph, tasked with delivering a final judgment on the truthfulness of a target declaration. 
It operates by first aggregating the complete body of evidence compiled throughout the verification process, which is sourced from all of its dependency nodes. 
This synthesized information is then presented to an LLM to perform a comprehensive analysis and render one of two possible labels: SUPPORTS if the evidence confirms the declaration otherwise REFUTES.

\subsection{Graph Execution Flow}
Each node in the DAG is pre-processed before execution to ensure the accuracy and consistency of the verification process. 
The process involves:
\begin{enumerate}
\item Evidence propagation: Evidence and output from parent nodes is propagated to the current node. This step ensures that all supporting information is available for task execution, enabling the node to operate on the latest and most relevant information.
\item Refine information Update: Check if one of the parent node is \texttt{REFINE} Node. If yes, we need to replace the input node with the output from that \texttt{REFINE} Node.
\end{enumerate}

The graph executes via topological sort. 
Execution begins with the root nodes running in parallel.
The system dynamically schedules a node for execution as soon as all its parent nodes have finished.
This process continues until all the nodes have been processed.

\subsection{Graph Modification for Dynamic Planner}
During the verification process, an agent might deemed a \texttt{THINK} Node or a \texttt{JUDGE} Node as unsolvable because of lacking information. 
A key feature of our framework is the dynamic generation of the verification actions. 
The core algorithm governing our dynamic tree generation and action execution is outlined in the Algorithm \ref{alg:dynamic-tree}.

\begin{algorithm}[t]
\small
\caption{Dynamic Tree Planner Algorithm}
\label{alg:dynamic-tree}
\begin{algorithmic}

\State \textbf{Input:} Claim ($\mathbf{C}$)
\State \textbf{Output:} $\mathbf{Label}$

\State Generate a Directed Acyclic Graph verification plan $\mathcal{G}$
\State Concurrently execute nodes in $\mathcal{G}$ following topological order

\For{each node $\mathbf{v} \in \mathcal{G}$}

    \State Let $\mathbf{v_{orig}}$ be the intrinsic input associated with node $v$
    \State Let $\mathrm{Parents}(v) = \{u \mid (u \rightarrow v) \in \mathcal{G}\}$

    \State Collect parent outputs:
    \[
        \mathbf{v_{parents}} = \{\mathrm{output}(u) : u \in \mathrm{Parents}(v)\}
    \]
    \State Combine intrinsic input and parent outputs:
    \[
        \mathbf{v_{input}} = \text{Combine}(\mathbf{v_{orig}}, \mathbf{v_{parents}})
    \]
    \If{type(v) = SEARCH}
        \State $\mathrm{output}(v) \gets \text{Search}(\mathbf{v_{input}})$

    \ElsIf{type(v) = REFINE}
        \State $\mathrm{output}(v) \gets \text{Refine}(\mathbf{v_{input}})$

    \ElsIf{type(v) = THINK}
        \State $\mathrm{output}(v) \gets \text{Reason}(\mathbf{v_{input}})$
        \If{evidence insufficient}
            \State Mark $v$ as failed
            \State Generate new subtree $\mathcal{G}_{sub}$ from $v$
            \State $G \gets G \cup \mathcal{G}_{sub}$
            \State Continue concurrent execution
        \EndIf

    \ElsIf{type(v) = JUDGE}
        \State $\mathrm{output}(v) \gets \text{Judge}(\mathbf{v_{input}})$
        \If{evidence insufficient}
            \State Mark $v$ as failed
            \State Generate new subtree $\mathcal{G}_{sub}$ from $v$
            \State $G \gets G \cup \mathcal{G}_{sub}$
            \State Continue concurrent execution
        \EndIf
    \EndIf

\EndFor

\State \textbf{Output:} Label = SUPPORTS / REFUTES

\end{algorithmic}
\end{algorithm}

For the \texttt{THINK} Node, Since this is a dependency tree, if a node is failing, we cannot perform accurate execution for their descendant. 
In such cases, the framework dynamically adapts by skipping over its descendant tasks and generating a new sub-tree $(\mathcal{G}_{sub})$ specifically designed to seek out the missing information. 
Once the new subtree is generated, it is integrated into the original graph $\mathcal{G}$. 
Ultimately, the \texttt{THINK} node outputs a finalized reasoning summary that is stored as its result, thereby consolidating its analytical step within the larger reasoning framework.
For example, in Figure \ref{fig:dynamic-tree}, after execution the thinking node, the output is evidence is insufficient, so our framework regenerate a new sub-tree to search more information.

For the \texttt{JUDGE} node, if the output is uncertain, the node does not simply exit but proactively triggers a graph correction to initiate a new subtree aimed at acquiring the missing evidence. Upon reaching a determinate verdict, the node's output consists of both the definitive label and a detailed textual explanation outlining the reasoning behind its final judgment.
This ensures that the verification process adapts dynamically to prevent the system from hallucinating or ending prematurely.
The complete pipeline is illustrated in Figure \ref{fig:dynamic-tree}.

\section{Experiment Setups}
\subsection{Datesets}
We evaluated our \texttt{Trification} framework using two widely adopted claim verification benchmarks: FEVEROUS \cite{Aly21Feverous} and HOVER \cite{jiang-etal-2020-hover}.
Both datasets consist of claims originating from Wikipedia and cover a diverse range of complexity that requires multi-hop reasoning.
Specifically, the HOVER dataset consists of 4,000 claims categorized into 3 complexity levels: 1,126 two-hop, 1,835 three-hop, and 1,039 four-hop claims.
The FEVEROUS dataset consist of 2,962 complex claims.
In line with the previous method \cite{pan2023factchecking}, we only select claims that require sentence-level evidence only.

\subsection{Baseline}
We compare \texttt{Trification} with various baselines that are categorized into four different categories:
\begin{itemize}
\item \textbf{Fine-tuned}: These methods leverage language models fine-tuned on claim verification dataset. Specifically, \texttt{BERT-FC} \cite{soleimani2019bertevidenceretrievalclaim} and \texttt{LisT5} \cite{jiang2021exploring} fine-tuned BERT and T5 models on the fact-checking benchmarks, respectively.
\texttt{RoBERTa-NLI} \cite{nie-etal-2020-adversarial} and \texttt{DeBERTa-NLI} \cite{he2021debertav3} leverage RoBERTa-large and DeBERTaV3 model and fine-tuned on FEVER \cite{thorne2018fever} dataset and four NLI datasets.
\texttt{MULTIVERS} \cite{wadden-etal-2022-multivers} uses a LongFormer \cite{beltagy2020longformer} model and tuned under FEVER dataset.
\item \textbf{LLM-based}: These methods perform claim verification by directly uses LLM through prompting technique.
This includes using zero-shot prompts using \texttt{ChatGPT} \cite{zhao-etal-2024-pacar} or 
few-shot prompts like \texttt{Codex} \cite{li2022evaluating} and \texttt{Flan-T5} \cite{chung2024scaling}.
\item \textbf{LLM Agent-based}: These methods leverage LLM as agents to execute specific sub-tasks within a verification pipeline.
\texttt{ProgramFC} \cite{pan2023factchecking} generates executable programs for operations like retrieval, question answering, or fact verification.
Similarly, \texttt{PACAR} \cite{zhao-etal-2024-pacar} introduces a multi-agent system that decomposes claim and assigns the resulting tasks to specialized tool executors, followed by a final verification step.
\item \textbf{LLM Agent-based with Dynamic Planning}: These method goes beyond standard LLM-agent based by adjusting their verification plan during execution.
\texttt{LoCaL} \cite{local} uses a multi-round decomposer for dynamic planning and two evaluating agents for two-way confidence updating to enhance logical and causal consistency. 
\texttt{SearChain} \cite{search-in-the-chain} uses LLM-based agents to dynamically verify and complete Chain-of-Query (CoQ) reasoning path, allowing it to iteratively correct and restructure its reasoning process.
\end{itemize}
\subsection{Implementation Details}
We employ a high performance model with zero-shot prompting technique as our primary LLM backbone.
For evaluation, we adopt the standard macro-f1 score \cite{pan2023factchecking, local, zhao-etal-2024-pacar} to assess claim verification performance.
We conduct experiments under two distinct settings:
\begin{itemize}
\item \textbf{Static Mode}: This is a baseline configuration designed for a fair comparison with other baselines such as \texttt{ProgramFC} \cite{pan2023factchecking}. 
In this mode, the verification graph is executed once without modification. The \texttt{SEARCH} node employs the Wiki strategy.
Specifically, the BM25 retriever from the Pyserini toolkit \cite{lin2021pyserini} over a static Wikipedia dump, fetching the top-10 paragraphs.
\item \textbf{Dynamic Mode}: This is the full expression of our framework's adaptive capability. 
Specifically, the \texttt{SEARCH} node employs the Search Engine strategy, fetching the top-10 relevant snippets.
The graph can undergo up to 3 modifications to recover from failures or refine its strategy.
\end{itemize}

Unless otherwise specified, results for our \texttt{Trification} framework are reported using the \textbf{Static Mode} to ensure a direct comparison with prior work that uses static knowledge bases.
Due to the computational and API costs associated with live search engine queries and iterative graph modifications, the results for the \textbf{Dynamic Mode} (presented in \autoref{subsection:dynamic-result}) are reported on a 150-claim subset of each test set.

\section{Experimental Results}
\subsection{Main Results}
\begin{table*}[]
    \small
    \centering
    \begin{tabular}{ll|ccc|c|c}
    \toprule
         \multicolumn{2}{l|}{\multirow{2}{*}{Models}} & \multicolumn{3}{c|}{HOVER} & \multirow{2}{*}{FEVEROUS} & \multirow{2}{*}{AVERAGE} \\
         & & 2-hop & 3-hop & 4-hop & & \\\midrule
         \multirow{5}{*}{I} & \texttt{BERT-FC} & 50.68 & 49.86 & 48.57 & 51.67 & 50.20  \\
         & \texttt{LisT5} & 52.56 & 51.86 & 50.46 & 54.15 & 52.27  \\
         & \texttt{RoBERTa-NLI} & 63.62 & 53.99 & 52.40 & 57.80 & 56.95  \\
         & \texttt{DeBERTaV3-NLI} & 68.72 & 60.76 & 56.00 & 58.81 & 61.07  \\
         & \texttt{MUTIVERS} & 60.17 & 52.55 & 51.86 & 56.61 & 55.30 \\ \hline

         \multirow{3}{*}{II} & \texttt{ChatGPT} & 66.94 & 60.56 & 58.73 & 55.72 & 60.49  \\
         & \texttt{Codex} & 65.07 & 56.63 & 57.27 & 62.58 & 60.39 \\
         & \texttt{Flan-T5} & 69.02 & 60.23 & 55.42 & 63.73 & 62.10  \\ \hline

         \multirow{3}{*}{III} & \texttt{ProgramFC (N=1)} & 69.36 & 60.63 & 59.16 & 67.80 & 64.24  \\
         & \texttt{ProgramFC (N=5)} & 70.30 & 63.43 & 57.74 & 68.06 & 64.88  \\
         & \texttt{PACAR} & \underline{73.13} & 64.07 & \underline{63.82} & \underline{72.61} & \underline{68.40}  \\ \hline

         \multirow{2}{*}{IV} & \texttt{LoCaL} & 72.71 & \underline{64.11} & 61.59 & 68.22 & 66.66  \\
         & \texttt{SearChain} &  64.46 & 60.30 & 56.54 & 66.69 & 62.00 \\ \hline
         
         Our & \texttt{Trification} & \textbf{75.13} & \textbf{66.42} & \textbf{66.23} & \textbf{74.72} & \textbf{70.63}  \\
        &  \texttt{w/o REFINE} & 71.22 & 61.56 & 60.99 & 71.30 & 66.26 \\
       &  \texttt{w/o REFINE and THINK} & 65.64 & 55.10 & 55.20 & 70.58 & 61.63 \\
         \bottomrule
    \end{tabular}
    \caption{Macro-F1 (\%) score on HOVER and FEVEROUS dataset. The best and second-best results in each column are indicated with bold and underlined text, respectively.}
    \label{tab:main-results}
\end{table*}

\autoref{tab:main-results} presents the macro F1-score of our \texttt{Trification} compared to state-of-the-art models across different categories.
On the HOVER dataset, our method achieves performance gains of 2.00\%, 2.35\%, and 2.41\% for the 2-hop, 3-hop, and 4-hop settings. 
The increasing trends in the improvements underscores the importance of how our tree-based planner and graph structure are crucial for managing the increased complexity of multi-hop claims.
Consequently, our framework successfully narrows the performance gap between complexity level, reducing the difference between 2-hop and 3-hop to to 8.71\%, and between 3-hop and 4-hop to just 0.19\%.
Similarly, our framework outperforms all state-of-the-art methods by 2.11\% on the FEVEROUS dataset.
In average, \texttt{Trification} improves fact-checking performance by 2.23\% across all evaluated benchmarks.

An interesting findings from the table is that LLM agents with dynamic planner underperform compared to agents with a fixed plan.
This suggests that unlimited dynamic planning can be detrimental without a guiding structure, potentially leading to unfocused and inefficient iteration over sub-claims.
Our framework addresses this weakness by generating a comprehensive set of actions and structuring them into a DAG, thereby providing the benefits of dynamic adaptation within a controlled, goal-oriented framework, which ultimately leads to superior performance.

\subsection{Results of \texttt{Trification} in Dynamic Settings}\label{subsection:dynamic-result}

Due to the significant computational overhead and API costs for performing live search engine queries and dynamic graph modifications, the evaluation of the \textbf{Dynamic Mode} was conducted on a curated subset of 150 challenging claims from each test set. 
These 'Hard' samples are defined as claims that the \textbf{Static Mode} could not verify correctly.

The results in \autoref{tab:dynamic-results} demonstrate that the our dynamic method substantially outperforms the static approach on this challenging subset.
This performance gain is more pronounced with the increasing claim complexity levels: 4-hop claims show a 50\% improvement, nearly double the 28\% gain for 2-hop claims (see \autoref{tab:dynamic-results-details}).
This outcome is achieved efficiently.
For the final results, we observe that the number of Uncertain predictions decreases under the dynamic tree setting, indicating that the dynamic strategy is effective. 
On average, our framework required 1.92 graph modifications on HOVER and 2.13 on FEVEROUS.
The number of \texttt{SEARCH} nodes increases slightly (2.74 → 3.51 on HOVER and 2.64 → 3.60 on FEVEROUS).
This indicates that our approach recovers from failures with minimal overhead, largely by using information from the original tree rather than initiating new searches for the same information.



Beyond search, the dynamic planner also introduces modest increases across other node types. The number of \texttt{THINK} nodes rises from 1.52 → 1.88 on HOVER and 1.67 → 2.16 on FEVEROUS, while \texttt{REFINE} nodes increase from 0.36 → 0.65 and 0.12 → 0.42, respectively. Likewise, \texttt{JUDGE} nodes grow from 1.09 → 1.53 on HOVER and 1.01 → 1.53 on FEVEROUS. These increases reflect the dynamic planner’s ability to explore alternative verification paths, but the magnitude remains modest—further supporting the conclusion that the method recovers from failures using minimal overhead, largely by reusing evidence from the original tree rather than repeatedly initiating new searches.

In terms of latency, the dynamic tree is approximately one second slower on average compared to the static tree. At the node level, \texttt{REFINE} nodes appear significantly more frequently. The \texttt{THINK} nodes in the dynamic tree also exhibit shorter average processing time than those in the static tree. A possible explanation is that dynamic-tree think nodes have a simpler objective—they only need to determine whether the existing evidence is sufficient—whereas static-tree think nodes must also generate intermediate stances and reasoning outputs. These findings collectively underscores the importance of of combining graph modification with a search engine to explore alternative verification paths for complex reasoning tasks.

\begin{table}[]
    \centering
    \resizebox{\columnwidth}{!}{%
        \begin{tabular}{lcc}
            \toprule
             & HOVER & FEVEROUS \\
            \midrule
            Graph Modification & 1.92\% & 2.13\% \\
            Macro-F1 & +40.0\% & +60.3\% \\
            \texttt{SEARCH} Node Increased & 2.74 $\rightarrow$ 3.51 & 2.64 $\rightarrow$ 3.60 \\
            \texttt{THINK} Node Increased & 1.52 $\rightarrow$ 1.88 & 1.67 $\rightarrow$ 2.16\\
            \texttt{REFINE} Node Increased & 0.36 $\rightarrow$ 0.65 & 0.12 $\rightarrow$ 0.42\\
            \texttt{JUDGE} Node Increased & 1.09 $\rightarrow$ 1.53 & 1.01 $\rightarrow$ 1.53 \\
            \bottomrule
        \end{tabular}%
    }
    \caption{Dynamic Tree Planner Average Performance on 150 hard samples from HOVER and FEVEROUS datasets. Hard samples are those our static planner cannot predict correctly.}
    \label{tab:dynamic-results}
\end{table}
Collectively, these findings highlight the effectiveness of combining graph modification with targeted search to explore alternative verification paths in complex, multi-hop reasoning tasks, achieving large performance gains with minimal computational overhead.


\subsection{Contribution of \texttt{REFINE} and \texttt{THINK} Nodes}
We examine the effectiveness of each modules in \texttt{Trification} by implementing several variants in the open book setting, as shown in the lower part of \autoref{tab:main-results}. The full model achieves the highest performance across all settings, with an average Macro-F1 score of 70.63\%, demonstrating the effectiveness of the complete reasoning pipeline that integrates \texttt{SEARCH}, \texttt{REFINE}, \texttt{THINK}, and \texttt{JUDGE} nodes. 
Removing the \texttt{REFINE} nodes leads to a consistent drop in performance (average = 66.26\%), indicating that iterative evidence refinement plays a crucial role in enhancing multi-hop reasoning accuracy. 
When both the \texttt{THINK} and \texttt{REFINE} nodes are excluded, performance declines further (average = 61.63\%), particularly on HOVER test data, confirming that the reasoning and refinement stages are vital for managing complex evidence trees. Overall, these results validate the complementary roles of the \texttt{REFINE} and \texttt{THINK} processes in enabling the model to generate coherent, evidence-grounded fact-checking decisions.


\subsection{Error Analysis}
To better understand the limitations of our system, we conducted a manual analysis on 10 failed claims. The analysis reveals two main sources of failure: (1) Search Engine Errors and (2) Ambiguous Claims. Table~\ref{tab:failure-analysis} presents representative examples of each failure type.

\paragraph{Search Engine 70\%}
Among the analyzed cases, 70\% (7/10) of the errors originated from the search engine component. Specifically, four cases (4/7) were due to the search engine retrieving evidence that \emph{contradicted} the ground truth, while the remaining three cases (3/7) occurred because the search engine \emph{failed to retrieve} relevant evidence.

In the first case type, the search results supported an incorrect interpretation of the claim. For example, the claim ``Gregg Rolie is not a keyboardist'' is labelled as Supports, but the search engine returned evidence asserting that ``Gregg Rolie is a keyboardist'' leading to a contradiction.  
In the second case type, the search engine could only retrieve partial evidence. For instance, the claim that ``Arthur Noss was a gunner at the Battle of Britain and Battle of Malta'' was partially supported—the retrieved evidence confirmed that Arthur Noss was a gunner, but not that he served in the two specified battles.

\paragraph{Ambiguous Claim 30\%}
The remaining 30\% (3/10) of failures resulted from ambiguities in the claim itself, which confused the reasoning process of either the planner or the judge module.  

For example, the claim “The by-election of a constituency represented in the House of Commons of the UK Parliament since 2010 by David Rutley was held on 30 September 1971” combines events from different time periods (2010 and 1971), making temporal interpretation unclear.  
Another example is the claim ``There were 331 episodes of the TV series where Julianna Margulies had the role of Carol Hathaway.'' The system misinterpreted this as implying that Julianna appeared in all 331 episodes, whereas the claim only asserts that she played the role in that TV series.

Developing post-processing strategies to handle conflicting search results or disambiguate unclear claims remains a challenge for future work.


\section{Conclusion}
We presented \texttt{Trification}, a novel automated fact-checking framework that ensures complete claim coverage through a comprehensive set of verification actions organized in a dependency graph. This graph models logical interactions and can be dynamically modified to adapt the verification strategy. Experiments on two challenging benchmarks show that \texttt{Trification} significantly improves fact-checking accuracy over previous methods. Future work includes handling conflicting search results and resolving ambiguities in claims to further enhance robustness and reliability.




\bibliography{custom}

\appendix

\section{Appendix}
\label{sec:appendix}

\subsection{Error Case Analysis Details}
\autoref{tab:failure-analysis} shows the error analysis for \texttt{Trification}.
\begin{table*}[t]
\centering
\begin{tabularx}{\textwidth}{>{\hsize=.2\hsize}X >{\hsize=.8\hsize}X}
\toprule
\textbf{Failure Type} & \textbf{Case Description} \\
\midrule
\textbf{Search Engine} & \textbf{Contradictory Evidence:} A Claim stated that Gregg Rolie is not a keyboardist is labelled as Supports. However, retrieved evidence asserted that he \emph{is} a keyboardist. \\
\midrule
\textbf{Search Engine} & \textbf{Incomplete Evidence:} Claim stated that Arthur Noss was a gunner at the Battle of Britain and Battle of Malta. The search engine only found evidence of him being a gunner, but not his participation in the two battles. \\
\midrule
\textbf{Ambiguous Claim} & \textbf{Temporal Ambiguity:} “The by-election of a constituency represented in the House of Commons of the UK Parliament since 2010 by David Rutley was held on 30 September 1971.” Ambiguous temporal relation between events in 1971 and 2010. \\
\midrule
\textbf{Ambiguous Claim} & \textbf{Quantitative Ambiguity:} “There were 331 episodes of the TV series where Julianna Margulies had the role of Carol Hathaway.” Misinterpreted as implying her appearance in all 331 episodes. \\
\bottomrule
\end{tabularx}
\caption{Representative failure cases from error analysis.}
\label{tab:failure-analysis}
\end{table*}


\subsection{Detail Results of 
\texttt{Trification} Dynamic Settings}
\autoref{tab:dynamic-results-details} presents the detailed performance results of dynamic \texttt{Trification}.
\begin{table*}[]
    \centering
    \begin{tabular}{l|ccc|c|c}
        \toprule
        \multirow{2}{*}{Models} & \multicolumn{3}{c|}{HOVER} & \multirow{2}{*}{FEVEROUS} & \multirow{2}{*}{AVERAGE} \\
         & 2-hop & 3-hop & 4-hop & & \\\midrule
         \texttt{Trification} & 00.00 & 00.00 & 00.00 & 00.00 & 00.00 \\
         \texttt{Trification-dynamic} & 28.00 & 42.00&50.00&60.26 & 45.07 \\
         \bottomrule
    \end{tabular}
    \caption{Performance on 150 hard samples of HOVER and FEVEROUS; Hard samples = Our static planner cannot predict this correctly}
    \label{tab:dynamic-results-details}
\end{table*}

\end{document}